\pdfoutput=1
\documentclass[11pt]{article}

\usepackage{acl}

\usepackage{times}
\usepackage{latexsym}

\usepackage[T1]{fontenc}
\usepackage[utf8]{inputenc}

\usepackage{microtype}

\usepackage{inconsolata}
\usepackage{graphicx}
\usepackage{pifont}
\usepackage{multirow}
\usepackage{courier}
\usepackage{booktabs}
\usepackage{hyperref}
\usepackage{amsmath}

\newcommand\modelname{PromptSLU}
\title{A Unified Framework for Multi-intent Spoken Language Understanding with prompting}

\author{Feifan Song, Lianzhe Huang and Houfeng Wang \\
        MOE Key Laboratory of Computational Linguistics, Peking University, China \\ 
        \texttt{songff@stu.pku.edu.cn}\\
        \texttt{\{hlz, wanghf\}@pku.edu.cn}
        }

\begin{document}
\maketitle

\begin{abstract}
Multi-intent Spoken Language Understanding has great potential for widespread implementation. Jointly modeling Intent Detection and Slot Filling in it provides a channel to exploit the correlation between intents and slots. However, current approaches are apt to formulate these two sub-tasks differently, which leads to two issues: 1) It hinders models from effective extraction of shared features. 2) Pretty complicated structures are involved to enhance expression ability while causing damage to the interpretability of frameworks. In this work, we describe a Prompt-based Spoken Language Understanding~({\modelname{}}) framework, to intuitively unify two sub-tasks into the same form by offering a common pre-trained Seq2Seq model. In detail, ID and SF are completed by concisely filling the utterance into task-specific prompt templates as input, and sharing output formats of key-value pairs sequence. Furthermore, variable intents are predicted first, then naturally embedded into prompts to guide slot-value pairs inference from a semantic perspective. Finally, we are inspired by prevalent multi-task learning to introduce an auxiliary sub-task, which helps to learn relationships among provided labels. Experiment results show that our framework outperforms several state-of-the-art baselines on two public datasets.
\end{abstract}

\section{Introduction}
\label{sec:intro}
Spoken Language Understanding~(SLU) is a fundamental part in the area of Task-oriented Dialogue~(TOD) modeling. It serves as the entrance of the pipeline with two sub-tasks: semantic information understanding by detecting user intents and extraction by filling the prepared slots, called Intent Detection~(ID) and Slot Filling~(SF), respectively. The former is usually modeled into a text classification problem and the latter is completed by cutting out fragments from an input utterance in the form of sequence tagging. 

\begin{figure}[t]
\centering
\includegraphics[width=0.48\textwidth]{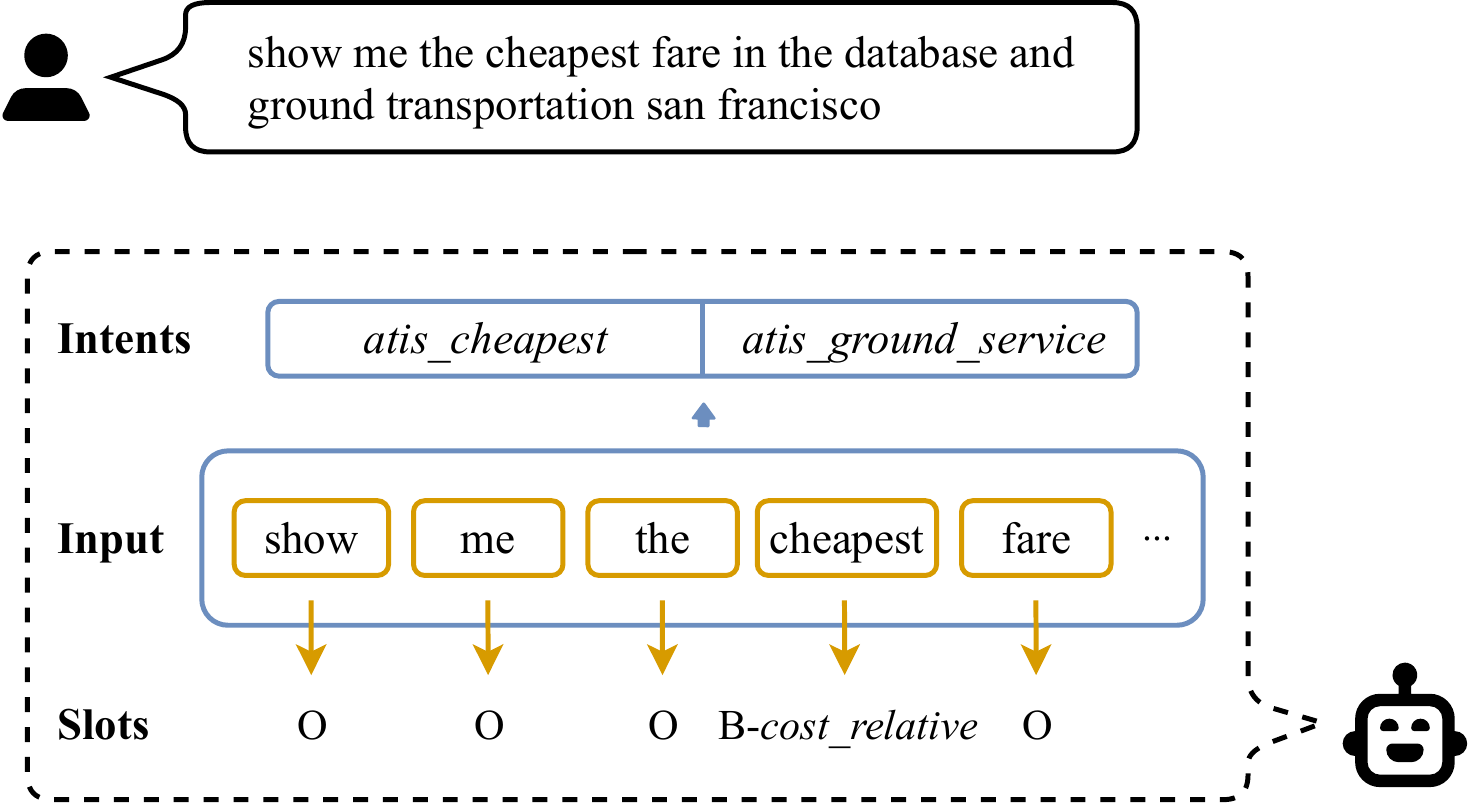}

\caption{Illustration of prior paradigm for jointly modelling multi-intent Spoken Language Understanding.}
\label{fig:intro} 
\end{figure}
Early works~\citep{ravuri2015recurrent,onlyID} focused on two sub-tasks separately, but the correlation between two sub-tasks is seen as the key to further improvement of SLU in recent years. Thus, such jointly modeling techniques were offered to bridge the gap of the common features between Intent Detection and Slot Filling. 
\citet{chen2019bert} treated BERT as a shared contextual encoder for two sub-tasks, while \citet{qin-etal-2019-stack} further brought token-level intent prediction results into Slot Filling process. Apart from these conventional NLU pathways, \citet{zhang2021joint} and \citet{tassias2021prompting} chose to formulate SLU as a constrained generation task.

However, real-world dialogue is more complex. Each utterance can be longer and contain more than one intent. As is shown in Figure~\ref{fig:intro}, the utterance ``Show me the cheapest fare in the database and ground transportation san francisco'' contains two distinct intents~(\textit{atis\_cheapest} and \textit{atis\_ground\_service}). For this kind of multi-intent SLU, similar jointly modelling methods are discussed~\citep{gangadharaiah-narayanaswamy-2019-joint,qin-etal-2020-agif,ijcai2021-523,qin-etal-2021-gl,9747843,9747477} and work well on several datasets, where multiple intents detection is often formulated as a multi-label text classification problem~\citep{gangadharaiah-narayanaswamy-2019-joint}, and potential interaction among intent and slot labels also plays an essential role~\citep{qin-etal-2020-agif,ijcai2021-523,qin-etal-2021-gl,9747843,9747477}.

Although following the path of jointly modeling, prior methods tend to integrate ID and SF during encoding by providing a shared feature extractor, but treat them individually when decoding. Figure~\ref{fig:intro} displays different decoding processes of two sub-tasks as multi-label classification and sequence tagging. The vast distinction in task formulation between ID and SF is indispensable. It discourages models from effective shared feature extraction and in turn, hinders the potential of comprehensively better performance. Consequently, unifying the two sub-tasks from start to finish with a common framework is significant. 

In this paper, we propose a framework to leverage pre-trained language models~(PLMs) and prompting to handle the aforementioned challenges, called \textbf{Prompt}-based \textbf{S}poken \textbf{L}anguage \textbf{U}nderstanding~({\modelname{}}) framework. Instead of using complicated structures to exploit the correlation among labels across two sub-tasks, our framework just incorporates both of them into text generation tasks with prompts, where the respective outputs share a general format of sequences of key-value pairs, namely Belief Span~\citep{lei-etal-2018-sequicity}. During inference, given an utterance and certain task requirement, {\modelname{}} first fills the utterance in a task-specific prompt template and then inputs it into a pre-trained Seq2Seq model.

Besides, consistency between two sub-tasks is crucial in SLU. Compared with prior settings, the multi-intent scenario is more challenging for accurate alignment from intents to slots because of the greater length of utterances and increased number of labels. For this issue, we explore an intuitive way in which intents are driven to restrain SF by plugging intents into prompt templates, namely Semantic Intents Guidance~(SIG). As is plain to humans in form, this design also allows the utilization of semantic information of intents to promote overall comprehension of our framework. Furthermore, inspired by multi-task learning used by \citet{paolini2020structured}, \citet{su-etal-2022-multi} and \citet{wang2022instructionner}, we try an auxiliary sub-task, called Slot Prediction~(SP), to steer models to additionally maintain semantic consistency. Experiments on two datasets demonstrate that {\modelname{}} outperforms SOTA baselines on most metrics, including those using PLMs. Abundant ablation studies also show the effectiveness of each component.

In summary, our contributions are as follows:
\begin{itemize}
\item We take the first step to introduce prompts into multi-intent SLU by transforming Intent Detection and Slot Filling into a common text generation formulation.
\item We present Semantic Intents Guidance Mechanism and a new sub-task Slot Prediction to provide more intuitive channels for interaction among texts, intents and slots. 
\item Experiments on two public datasets show better results of {\modelname{}} compared with methods with SOTA performance, and the effectiveness of proposed strategies and semantic information.
\end{itemize}

\section{Related Work}
\label{sec:related}
\subsection{Spoken Language Understanding}
Spoken Language Understanding~(SLU), in the task-oriented dialog system, is mainly composed of two sub-tasks, Intent Detection~(ID) and Slot Filling~(SF). As for task formulations, most existing methods model ID into a text classification problem and SF into a sequence tagging problem. Early works separately handled ID and SF~\citep{schapire2000boostexter,1198860,raymond2007generative,5947636,ravuri2015recurrent,onlyID,wang2021enhanced}. However, current approaches prefer to modeling them together, with the consideration of the high correlation between them, and thus lead to substantial improvement. Jointly modeling techniques basically correspond to two different methodologies: 

\paragraph{Parallel Model} where ID and SF get outputs respectively, but share an utterance encoder, in an attempt to exploit the common latent features~\citep{zhang2016joint,hakkani-tr2016multi-domain,zhang2019using}. 
\paragraph{Serial Model} built by detecting intent in the first place and then utilizing intent information to navigate SF~\citep{zhang-etal-2019-joint,qin-etal-2019-stack,tassias2021prompting}. 

Our framework takes both sides mentioned above. First, It can be serially implemented according to the second methodology, i.e., allowing intents to benefit SF. Then, the results of two sub-tasks come from the same Seq2Seq model in this framework.

\subsection{Multi-intent SLU}
Multi-intent utterances tend to appear, at a higher frequency in reality, than those with a single intent. This problem has an increasing popularity in society~\citep{onlyID,gangadharaiah-narayanaswamy-2019-joint,qin-etal-2020-agif,ijcai2021-523,qin-etal-2021-gl,9747843,9747477}. \citet{gangadharaiah-narayanaswamy-2019-joint} used an attention-based neural network for jointly modeling ID and SF in multi-intent scenarios. \citet{qin-etal-2020-agif} proposed AGIF that organizes intent labels and slot labels together in the form of graph structure and focuses on the interaction of labels. \citet{ijcai2021-523} and \citet{qin-etal-2021-gl} utilize graph attention network to attend on the association among labels. Despite the attempt to build the bridge between intent and slot labels, ID and SF are still modeled in different forms, i.e., multi-label classification for ID but sequence tagging for SF. The essential problem has not been handled that distinct modeling ways hinder the extraction of common features. Our text-generation-based framework differs from the above approaches by unifying the formulation of ID and SF. On one hand, this way is plain to humans and naturally suitable for ID with variable intents. On the other hand, every step of promotion of our framework effectively benefits both sub-tasks. Our framework takes both sides above to be equipped with explicitly hierarchical implementation, while the results of two sub-tasks come from the same Seq2Seq model.

\subsection{Prompting}
GPT-3~\citep{NEURIPS2020_1457c0d6} provides a novel insight on the area of Natural Language Processing with the power of new paradigm~\citep{liu2021pre}, namely prompt. Currently, it has been explored in many directions by transforming the original task into a text-to-text form~\citep{zhong-etal-2021-adapting-language,lester-etal-2021-power,cui-etal-2021-template,wang2022instructionner,tassias2021prompting,su-etal-2022-multi}, where fine-tuning on downstream tasks effectively refers to knowledge from long-term pre-training processes, and especially works well on low-resource scenarios. \citet{zhong-etal-2021-adapting-language} tried to align the form of fine-tuning to next word prediction to optimize performance on text classification tasks. In another way, \citet{lester-etal-2021-power} appealed to soft-prompt to capture implicit and continuous signals. For sequence tagging, prompts appear as templates for text filling or natural language descriptions of instruction~\citep{cui-etal-2021-template,wang2022instructionner}. As to SLU in task-oriented dialogue~(TOD), it consists of two sub-tasks ID and SF which have homologous forms. \citet{tassias2021prompting} depended on prior distributions of intents and slots to generate a prompt containing all slots to be prdedicted before inference, lowering the difficulty of SF. \citet{su-etal-2022-multi} proposed a prompt-based framework PPTOD to integrate several parts in the pipeline of TOD, including single-intent detection that is different from our setting. However, such jointly modeling fashions neglect the complexity of co-occurrence and semantic similarity among tokens, intents and slots in multi-intent SLU, consequently suppressing the potential of PLMs. To this end, we build the Semantic Intents Guidance mechanism and design an auxiliary sub-task Slot Prediction, to exploit more detailed generality together.

\begin{figure*}[t]
\centering
\includegraphics[width=1.0\textwidth]{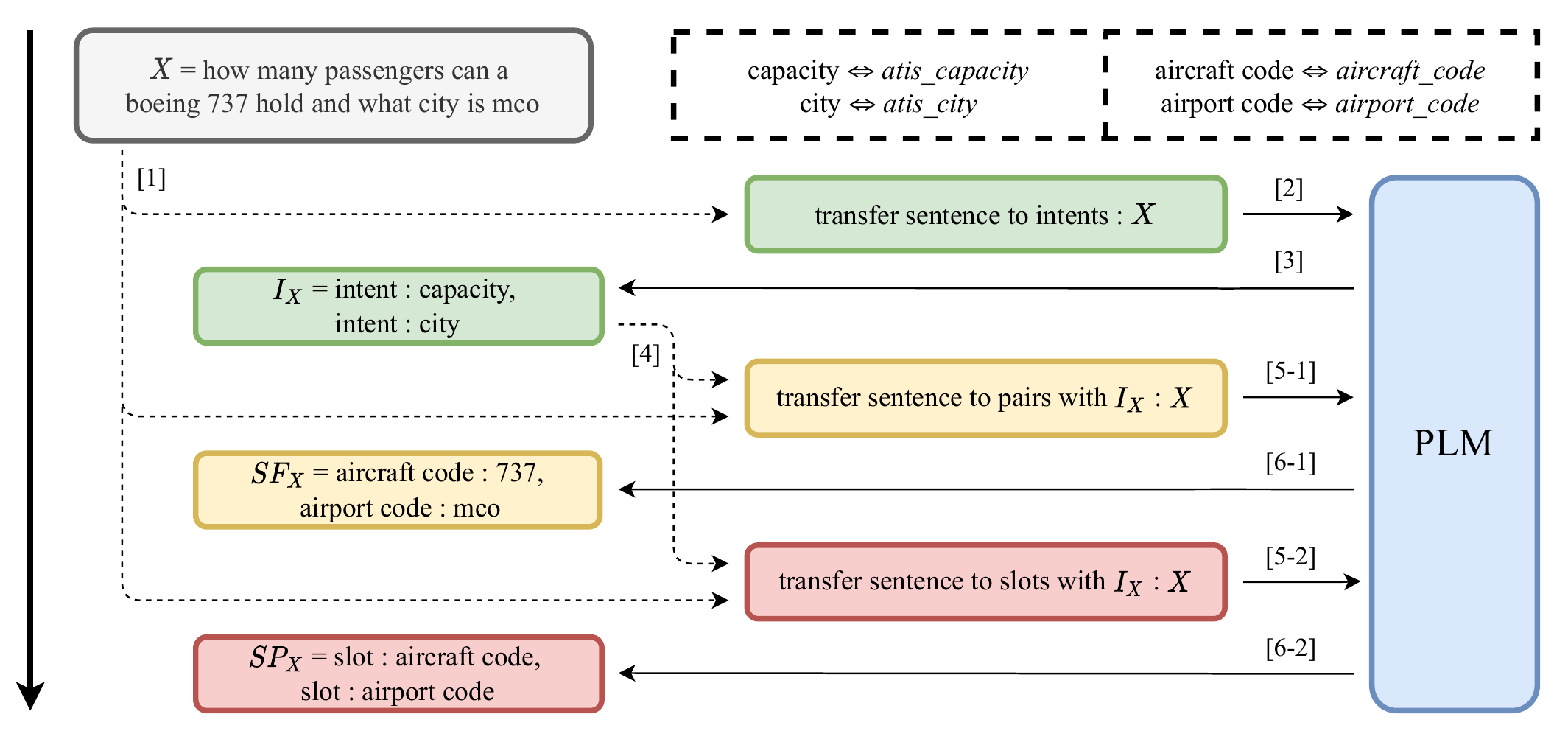}

\caption{\textbf{The flow path of \modelname{}.} We unify the formulations of each sub-task with a sharing PLM. Identical equations in the dashed rectangles present transformations between intent/slot labels and corresponding descriptions. The left arrow shows the direction of the flow path. [1] represents that input utterance is embedded into prompt templates for different sub-tasks. [2] and [3] are input and output of Intent Detection. [4] is the Semantic Intent Guidance mechanism where predicted intents are embedded into prompt templates to guide other sub-tasks during inference. [5-1] and [5-2] are inputs of Slot Filling and Slot Prediction, respectively, both of which can be executed in parallel. [6-1] and [6-2] are their outputs.}
\label{fig:overview} 
\end{figure*}

\section{Methodology}
\label{sec:methodology}
In this section, we describe the proposed {\modelname{}} along the inner flow of data. In this framework, we utilize different prompts to complete different sub-tasks, which is an intuitive way similar to \citet{su-etal-2022-multi}. However, {\modelname{}} is specially designed for multi-intent SLU that contains a particular intent-slot interaction mechanism, namely \textbf{Semantic Intent Guidance}~(SIG). We also propose a new auxiliary sub-task \textbf{Slot Prediction}~(SP) to improve Intent Detection~(ID) and Slot Filling~(SF).

Given the utterance $X = \{x_1, x_2, x_3, ..., x_N \}$, for each of sub-tasks including ID, SF and SP, {\modelname{}} inputs to the backbone model the concatenation of a task-specific prefix and X, then makes predictions. During SF or SP, intents can also be involved. The whole framework is shown in Figure~\ref{fig:overview}. Three sub-tasks are jointly trained with a sharing pre-trained language model~(PLM). 

\subsection{Intent Detection}
Traditionally, intents are defined as a fixed number of labels. A model is fed with input text, then maps it to these labels. Differently, we model the task in the form of text generation, i.e., the backbone model produces a sequence of intents corresponding to the input utterance.

Belief Span is first utilized by Sequicity~\citep{lei-etal-2018-sequicity} to cover filled and requestable slots for state tracking. We transfer this structure to the area of multi-intent SLU by defining the format of output sequence as:
\begin{equation}
    I_{X} = \textrm{intent : }i_1,..., \textrm{intent : }i_{N_{I}}
\end{equation}
where $i_m~(1\leq m\leq N_{I})$ is the $m$-th intent and $N_{I}$ is the number of predicted intents. As is clarified before, we integrate the task modeling by imposing similar formats on the following sub-tasks. The PLM takes the prompt ``\texttt{transfer sentence to intents : $X$}'', then carries out Intent Detection. 

Like tricks used in \citet{tassias2021prompting} and \citet{wang2022instructionner}, we use natural language descriptions to represent non-semantic intent labels, reducing the difficulty of exploiting semantic information. We also apply this operation in the follow-up sub-tasks, based on the assumption that {\modelname{}} significantly relies on semantic information to complete multi-intent SLU.

\subsection{Semantic Intent Guidance}
It is noted that there exists semantic similarity among tokens, intents and slots. This property is reflected in the aspects of not only frequent co-occurrence but also semantic resemblance, which may be of help for inference. We also consider it in our framework. 

We adopt the idea of jointly modeling by parsing intents from the output sequence of ID, and using them to facilitate other sub-tasks. Specifically, together with task-specific prefixes, intents also serve as an essential part of prompts for SF and SP in our prompt-based framework, to guide their completion and keep semantic consistency between ID and SF. We name it as the \textbf{Semantic Intent Guidance}~(SIG) mechanism and display it in some prompt templates later. 

\subsection{Slot Filling}
To let PLM generate slot-value pairs directly, raw data have to be processed in advance. We first extract golden slot-value pairs from each BIO-tagged sequence, based on tagging rules. Then we orderly assemble these golden pairs into one sentence in the format of Belief Span, which serves as text-generation target $SF_{x}$. This step is crucial for purpose of integrating task formulations:
\begin{equation}
    SF_{X} = s_1\textrm{ : }v_1,...,s_{N_{P}}\textrm{ : }v_{N_{P}}
\end{equation}
where $N_{P}$ denotes the number of predicted slot-value pairs.

Similar to the mapping procedure in Intent Detection, we also replace slot labels with corresponding natural language descriptions. The dashed boxes on the top of Figure~\ref{fig:overview} demonstrate the transformation from intent and slot labels to more human-readable phrases. This procedure occurs both in input and target utterances.

We introduce the SIG mechanism here, as mentioned above, to allow predicted intents to act as pilots for slot filling. In detail, we plug the intent phrase and user utterance into the task-specific prompt template: ``\texttt{transfer sentence to pairs with $I_X$ : $X$}''. We also do this in the next sub-task.

\subsection{Slot Prediction}
Multi-task learning is apt to offer a channel for the interaction of different tasks, and enhance supervision signals towards training objectives. Following multi-task settings in \citet{su-etal-2022-multi} and \citet{wang2022instructionner}, we design this auxiliary sub-task in the training stage, forcing the PLM to focus on the semantic interaction among tokens, intents and slots. We also leverage it for the improvement of maintaining semantic consistency. {\modelname{}} is required to generate slots sequence:
\begin{equation}
    SP_{X} = \textrm{slot : }s_1,..., \textrm{slot : }s_{N_{S}}
\end{equation}
according to the prefix ``\texttt{transfer sentence to slots with $I_X$ : $X$}''. We use $N_{S}$ to denote the number of predicted slots.

\subsection{Training}
Similar to other text-to-text tasks, our framework is trained to minimize negative log-likelihood for each sub-task:
\begin{equation}
    -\sum_{i=1}^{\left | Y \right |} \log_{}{p_{\Theta}\left(y_i\middle|y_{<i},X\right)} 
\end{equation}
where $Y$ denotes the target token sequence and $\Theta$ denotes model parameters. We introduce two variants of loss functions.

\textbf{Weighted Loss} integrates three sub-tasks by calculating a weighted sum for each sample:
\begin{equation}
    \mathcal{L}_w = \alpha\mathcal{L}_{ID} + \beta\mathcal{L}_{SP} + \gamma\mathcal{L}_{SF} 
\end{equation}
where $\alpha$, $\beta$ and $\gamma$ are hyper-parameters. $\mathcal{L}_{ID}$, $\mathcal{L}_{SP}$ and $\mathcal{L}_{SF}$ represent the loss of Intent Detection, Slot Prediction and Slot Filling, respectively. 

\textbf{Split Loss} is utilized for a shuffled dataset, which is built by splitting each sample in the original dataset into three text generation samples corresponding to three original labels for different sub-tasks. That is,
\begin{equation}
    \mathcal{L}_s(X,Y) =
    \begin{cases}
        \mathcal{L}_{ID}(X,Y)& {Y \in ID}\\
        \mathcal{L}_{SP}(X,Y)& {Y \in SP}\\
        \mathcal{L}_{SF}(X,Y)& {Y \in SF}
    \end{cases}
\end{equation}

\section{Experiment}
\label{sec:experiment}

\subsection{Datasets and Metrics}

We conduct experiments mainly on \textbf{MixATIS} and \textbf{MixSNIPS}~\citep{qin-etal-2020-agif}, which are widely used multi-intent SLU datasets. There are 13162, 759 and 828 samples in MixATIS for training, validation and test, respectively, as well as 39776, 2198 and 2199 in MixSNIPS.

Following \citet{9747477} and \citet{qin-etal-2021-gl}, we evaluate the performance of Slot Filling with F1 score, Intent Detection and sentence-level semantic frame parsing with accuracy, named \textbf{Slot F1}\footnote{Considering there exist samples with repeated slot-value pairs, we calculate Slot F1 after the elimination of duplicate pairs for {\modelname{}}.}, \textbf{Intent Acc} and \textbf{Overall Acc}.

\subsection{Baselines}
For evaluation of multi-intent SLU, we compare our approach with three baselines and their promoted versions~(if provided):\\
\textbf{AGIF}~\citep{qin-etal-2020-agif} : This method utilizes a graph interaction module to build token-wise connections between slot tags and intent labels.\\
\textbf{GL-GIN}~\citep{qin-etal-2021-gl} : Different from AGIF, this work exploits local coherence during slot tagging and sentence-level connections between intents and slots. It performs decoding in a non-auto-regressive way. GL-GIN-RoBERTa displaces self-attentive encoder by RoBERTa$_{\scriptsize \textrm{base}}$~(128M parameters)~\citep{https://doi.org/10.48550/arxiv.1907.11692}.\\
\textbf{SDJN}~\citep{9747843} : This work reformulates multi-intent detection as a weakly supervised task and design a self-distillation mechanism to circularly refresh the pipeline. SDJN-BERT likewise displaces the self-attentive encoder by BERT, whose size is not mentioned in the source paper.

\begin{table*}
\centering
\scalebox{0.78}{
  \begin{tabular}{llllllll}
    \toprule
    \multicolumn{2}{l}{\multirow{2}{*}{\textbf{Methods}}} & \multicolumn{3}{c}{\textbf{MixATIS}} & \multicolumn{3}{c}{\textbf{MixSNIPS}} \\ \cmidrule(lr){3-5}\cmidrule(lr){6-8}
    \multicolumn{2}{l}{}                                   & \multicolumn{1}{c}{Slot(F1)} & \multicolumn{1}{c}{Intent(Acc)} & \multicolumn{1}{c}{Overall(Acc)} & \multicolumn{1}{c}{Slot(F1)} & \multicolumn{1}{c}{Intent(Acc)} & \multicolumn{1}{c}{Overall(Acc)} \\ \midrule
    \multicolumn{1}{l}{\multirow{3}{*}{Non-pre-trained}} & \multicolumn{1}{l}{AGIF~\citep{qin-etal-2020-agif}} & \multicolumn{1}{c}{86.7} & \multicolumn{1}{c}{74.4} & \multicolumn{1}{c}{40.8} & \multicolumn{1}{c}{94.2} & \multicolumn{1}{c}{95.1} & \multicolumn{1}{c}{74.2} \\
    \multicolumn{1}{l}{}                                 & \multicolumn{1}{l}{GL-GIN~\citep{qin-etal-2021-gl}} & \multicolumn{1}{c}{88.3} & \multicolumn{1}{c}{76.3} & \multicolumn{1}{c}{43.5} & \multicolumn{1}{c}{94.9} & \multicolumn{1}{c}{95.6} & \multicolumn{1}{c}{75.4} \\
    \multicolumn{1}{l}{}                                 & \multicolumn{1}{l}{SDJN~\citep{9747843}} & \multicolumn{1}{c}{88.2} & \multicolumn{1}{c}{77.1} & \multicolumn{1}{c}{44.6} & \multicolumn{1}{c}{94.4} & \multicolumn{1}{c}{96.5} & \multicolumn{1}{c}{75.7} \\ \midrule
    \multicolumn{1}{l}{\multirow{4}{*}{Pre-trained}}     & \multicolumn{1}{l}{{SDJN+BERT}} & \multicolumn{1}{c}{87.5} & \multicolumn{1}{c}{78.0} & \multicolumn{1}{c}{46.3} & \multicolumn{1}{c}{95.4} & \multicolumn{1}{c}{96.7} & \multicolumn{1}{c}{79.3} \\

    \multicolumn{1}{l}{}                                 & \multicolumn{1}{l}{{GL-GIN+RoBERTa}} & \multicolumn{1}{c}{88.6} & \multicolumn{1}{c}{79.2} & \multicolumn{1}{c}{53.9} & \multicolumn{1}{c}{96.0} & \multicolumn{1}{c}{97.4} & \multicolumn{1}{c}{82.4} \\ \cmidrule{2-8}
    \multicolumn{1}{l}{}                                 & \multicolumn{1}{l}{{\modelname{}}$_{\mathcal{L}_{w}}$} & \multicolumn{1}{c}{88.6} & \multicolumn{1}{c}{83.6} & \multicolumn{1}{c}{53.3} & \multicolumn{1}{c}{96.2} & \multicolumn{1}{c}{\textbf{97.7}} & \multicolumn{1}{c}{82.8} \\
    \multicolumn{1}{l}{}                                 & \multicolumn{1}{l}{{\modelname{}}$_{\mathcal{L}_{s}}$} & \multicolumn{1}{c}{\textbf{89.6}*} & \multicolumn{1}{c}{\textbf{85.8}*} & \multicolumn{1}{c}{\textbf{57.2}*} & \multicolumn{1}{c}{\textbf{96.5}*} & \multicolumn{1}{c}{97.5} & \multicolumn{1}{c}{\textbf{84.8}*} \\
    \bottomrule
  \end{tabular}
}
\caption{\label{main_results}
    Main results on MixATIS and MixSNIPS. Methods are distinguished by whether pre-trained language models are involved. * means that the improvement beyond SOTA is significant with $p<0.05$ under t-test.
}
\end{table*}

For baselines except GL-GIN+RoBERTa, we directly take the results from the literature.

\subsection{Implementation Detail}
In this work, we choose T5~\citep{raffel2019exploring} as the backbone of {\modelname{}} with small~(60M parameters) and base~(220M parameters) versions, because our prompt design is coherent with the pre-trained tasks of T5. We set epoch and batch size to 10 and 16 on each processor, respectively. The length of sequences and the learning rate are uniformly set to 128 and 1e-4, while the dropout rate is 0.1 for MixATIS and 0.15 for MixSNIPS. We empirically set $\alpha$, $\beta$ and $\gamma$ as 1.0, 2.0 and 1.0, respectively. All test results are selected according to the performance of Overall Acc on the validation set. We report the average scores after running the code 4 times repeatedly with different global random seeds 0, 1, 2 and 3. Following \citet{su-etal-2022-multi}, our framework is built on Huggingface Library~\citep{wolf-etal-2020-transformers}. The optimizer here remains AdamW~\citep{loshchilov2018decoupled}. We conduct experiments on a Linux server with Intel Xeon E5-2680 and Nvidia GeForce RTX 3090. 

\subsection{Main Results}
Table~\ref{main_results} displays the experiment results of our framework and baselines. It is observed that our framework outperforms each baseline on each metric for both MixATIS and MixSNIPS, no matter whether a pre-trained model is leveraged. 

It is clear that pre-trained models are beneficial to this task. In general, methods with pre-trained models have better performance than methods without them. It mainly comes from the utilization of knowledge contained. Among them, {\modelname{}} is distinguished from other NLU methods by more concise modeling but better performance. This result verifies the efficient knowledge exploitation of the proposed prompt-based text generation paradigm for multi-intent SLU.

More specifically, when compared with SOTA results, \modelname{}$_{\mathcal{L}_{s}}$ still accomplishes significant advances. It should be noted that the amazing result on Overall Acc, which should have been limited by the relative slight improvement on Slot F1, is in fact beyond the SOTA result by a large margin. This indicates that during inference, our framework is skilled at maintaining semantic consistency between intents and slots in the same utterance. We preliminarily attribute this ability to the influence of the proposed Semantic Intent Guidance mechanism and Slot Prediction. 

We compare the performance between \modelname{}$_{\mathcal{L}_{w}}$ and \modelname{}$_{\mathcal{L}_{s}}$. Obviously, \modelname{}$_{\mathcal{L}_{s}}$ beats \modelname{}$_{\mathcal{L}_{w}}$ on almost all metrics, except close value of Intent Acc for MixSNIPS. It primarily demonstrates the superiority of ${\mathcal{L}_{s}}$ against the straightforward weighted sum of losses. We provide a careful analysis at Sec~\ref{ablation}. For other experiments, we use ${\mathcal{L}_{s}}$ as the loss function by default. 

\begin{table*}
\centering
\scalebox{0.78}{
\begin{tabular}{llllllll}
\toprule
\multicolumn{2}{l}{\multirow{2}{*}{\textbf{Methods}}} & \multicolumn{3}{c}{\textbf{MixATIS}} & \multicolumn{3}{c}{\textbf{MixSNIPS}} \\ \cmidrule(lr){3-5}\cmidrule(lr){6-8}
\multicolumn{2}{l}{}  & \multicolumn{1}{c}{Slot(F1)} & \multicolumn{1}{c}{Intent(Acc)} & \multicolumn{1}{c}{Overall(Acc)} & \multicolumn{1}{c}{Slot(F1)} & \multicolumn{1}{c}{Intent(Acc)} & \multicolumn{1}{c}{Overall(Acc)} \\ \midrule
\multicolumn{1}{l}{\multirow{10}{*}{w/o golden intents}} & \multicolumn{1}{l}{{\modelname{}}$_{\scriptsize\textrm{small}}$} & \multicolumn{1}{c}{87.1} & \multicolumn{1}{c}{80.4} & \multicolumn{1}{c}{50.6} & \multicolumn{1}{c}{95.4} & \multicolumn{1}{c}{97.4} & \multicolumn{1}{c}{79.4} \\ 
\multicolumn{1}{l}{} & \multicolumn{1}{l}{{\modelname{}}$_{\scriptsize\textrm{base}}$} & \multicolumn{1}{c}{\textbf{89.6}} & \multicolumn{1}{c}{\textbf{85.8}} & \multicolumn{1}{c}{\textbf{57.2}} & \multicolumn{1}{c}{\textbf{96.5}} & \multicolumn{1}{c}{97.5} & \multicolumn{1}{c}{\textbf{84.8}} \\
\multicolumn{1}{l}{} & \multicolumn{1}{l}{\quad $-$SP} & \multicolumn{1}{c}{87.9} & \multicolumn{1}{c}{82.6} & \multicolumn{1}{c}{50.8} & \multicolumn{1}{c}{96.2} & \multicolumn{1}{c}{97.3} & \multicolumn{1}{c}{82.9} \\ 
\multicolumn{1}{l}{} & \multicolumn{1}{l}{\quad $-$SIG} & \multicolumn{1}{c}{88.8} & \multicolumn{1}{c}{84.6} & \multicolumn{1}{c}{53.3} & \multicolumn{1}{c}{96.2} & \multicolumn{1}{c}{97.5} & \multicolumn{1}{c}{82.7} \\ 
\multicolumn{1}{l}{} & \multicolumn{1}{l}{\quad only ID} & \multicolumn{1}{c}{-} & \multicolumn{1}{c}{81.4} & \multicolumn{1}{c}{-} & \multicolumn{1}{c}{-} & \multicolumn{1}{c}{97.0} & \multicolumn{1}{c}{-} \\ 
\multicolumn{1}{l}{} & \multicolumn{1}{l}{\quad only SF} & \multicolumn{1}{c}{88.6} & \multicolumn{1}{c}{-} & \multicolumn{1}{c}{-} & \multicolumn{1}{c}{96.1} & \multicolumn{1}{c}{-} & \multicolumn{1}{c}{-} \\
\multicolumn{1}{l}{} & \multicolumn{1}{l}{\quad $\mathcal{L}_{w}^{(1,1,1)}$} & \multicolumn{1}{c}{88.3} & \multicolumn{1}{c}{81.5} & \multicolumn{1}{c}{51.7} & \multicolumn{1}{c}{96.3} & \multicolumn{1}{c}{97.3} & \multicolumn{1}{c}{83.3} \\ 
\multicolumn{1}{l}{} & \multicolumn{1}{l}{\quad $\mathcal{L}_{w}^{(2,1,1)}$} & \multicolumn{1}{c}{89.0} & \multicolumn{1}{c}{80.2} & \multicolumn{1}{c}{52.6} & \multicolumn{1}{c}{96.3} & \multicolumn{1}{c}{97.6} & \multicolumn{1}{c}{83.1} \\ 
\multicolumn{1}{l}{} & \multicolumn{1}{l}{\quad $\mathcal{L}_{w}^{(1,2,1)}$} & \multicolumn{1}{c}{88.6} & \multicolumn{1}{c}{83.6} & \multicolumn{1}{c}{53.3} & \multicolumn{1}{c}{96.2} & \multicolumn{1}{c}{\textbf{97.7}} & \multicolumn{1}{c}{82.8} \\ 
\multicolumn{1}{l}{} & \multicolumn{1}{l}{\quad $\mathcal{L}_{w}^{(1,1,2)}$} & \multicolumn{1}{c}{88.3} & \multicolumn{1}{c}{81.5} & \multicolumn{1}{c}{51.0} & \multicolumn{1}{c}{96.4} & \multicolumn{1}{c}{97.5} & \multicolumn{1}{c}{83.9} \\ 
\midrule
\multicolumn{1}{l}{\multirow{2}{*}{w/ golden intents}} & \multicolumn{1}{l}{{\modelname{}}$_{\scriptsize\textrm{small}}$} & \multicolumn{1}{c}{87.7} & \multicolumn{1}{c}{80.4} & \multicolumn{1}{c}{50.6} & \multicolumn{1}{c}{95.8} & \multicolumn{1}{c}{97.4} & \multicolumn{1}{c}{79.4} \\
\multicolumn{1}{l}{} & \multicolumn{1}{l}{{\modelname{}}$_{\scriptsize\textrm{base}}$} & \multicolumn{1}{c}{89.7} & \multicolumn{1}{c}{85.8} & \multicolumn{1}{c}{57.2} & \multicolumn{1}{c}{97.0} & \multicolumn{1}{c}{97.5} & \multicolumn{1}{c}{84.8} \\
\bottomrule
\end{tabular}
}
\caption{\label{ablation_results}
Ablation experiments. SP and SIG denote Slot Prediction and Semantic Intent Guidance, respectively. $\mathcal{L}_{w}^{(a,b,c)}$ represents that \modelname{}$_{\scriptsize\textrm{base}}$ is trained under the weighted loss function with ${(\alpha,\beta,\gamma )}$ set to ${(a,b,c)}$.
}
\end{table*}

\subsection{Ablation Study}
\label{ablation}
In this part, we abstract several factors and analyze their effectiveness. Results are shown in Table~\ref{ablation_results}. The reason why the improvement is relatively slight for MixSNIPS is that features and patterns in it are more straightforward to capture to some extent, thus encouraging each variant to have competitive performance. An intuitive idea is that model size plays a dominant factor in model performance, which is proved by the comparison between \modelname{}$_{\scriptsize\textrm{small}}$ and \modelname{}$_{\scriptsize\textrm{base}}$. Here \modelname{}$_{\scriptsize\textrm{small}}$ has a competitive performance compared to baselines with pre-trained models, while the quantity of parameters in it is pretty less. However, with expanding framework scale of backbone, \modelname{}$_{\scriptsize\textrm{base}}$ makes large improvements against \modelname{}$_{\scriptsize\textrm{small}}$.

We next remove Slot Prediction~(SP) while retaining Intent Detection and Slot Filling. We can observe that there are drops on every metric no matter which dataset, especially on Overall Acc. It confirms that SP serves as a catalyst to help maintain semantic consistency between ID and SF. 

We then check the effect of the SIG mechanism by changing the task-specific prompt templates and removing the blanks originally prepared for predicted intents. This also brings out performance decline. Similar to the ablation of SP, SIG also facilitates the consistency between two kinds of labels, which is one of the chief strengths of jointly modeling. We surprisingly observe that score of Intent Acc also decreases on MixATIS. We attribute it to some implicit semantic supervision during the update of model parameters, i.e., although parameters gradient cannot be propagated across sub-tasks, our framework still has the potential to revise intents prediction against SIG.

In {\modelname{}}, we implement jointly modeling for SLU along two directions. One is to unify two sub-tasks in a shared formulation with a common Seq2Seq model, as the other allows predicted intents to assist Slot Filling. Hence, we split two sub-tasks by accomplishing each with an exclusive T5 model, in order to evaluate the impact of jointly modeling. It is interesting that {\modelname{}} with only SF has competitive performance on Slot F1 when compared with other jointly modeling versions. However, the performance of {\modelname{}} with only ID has dropped a lot on Intent Acc for MixATIS. It confirms the effectiveness of the jointly modeling paradigm even in text generation formulation.

We propose two kinds of loss functions for \modelname{}. The weighted ${\mathcal{L}_{w}}$ has a similar structure to those used in baselines, while ${\mathcal{L}_{s}}$ refers to each sub-tasks as a text generation process more directly. Here we try different combinations of weight hyper-parameters for ${\mathcal{L}_{w}}$, but ${\mathcal{L}_{s}}$ still derives better performance than them in general. We assume that the optimization targets of different sub-tasks tend to be not identical, thus a trade-off of these targets is indispensable. Different weight hyper-parameters suggest the game, but the desired combination is likely to be hard to access, even dynamically changing along the training process. In contrast, ${\mathcal{L}_{s}}$ allows \modelname{} to avoid the difficulty of trade-off in each turn, but focus on completing just the text generation task and achieve better improvements.

During inference, \modelname{} with SIG generates slot-value pairs based on the predicted intents, where transmission errors should be considered. We experimentally estimate their effect on Slot Filling, by replacing predicted intents with the golden ones. The last two rows in Table~\ref{ablation_results} give the results. It can be seen that \modelname{} with golden intents generally has extremely slight improvement on Slot F1, while no increase in the values of Overall Acc. It indicates the effect of transmission errors is not significant. We mainly attribute it to the fact that \modelname{} checks those intents against the input utterance, and filters wrong and missing intents in the prompts, which proves that \modelname{} learns to understand semantics. Another reason may be the relatively high values of Intent Acc that reduce the impact of transmission errors.

\begin{figure}[t]
\centering
\includegraphics[width=0.38\textwidth]{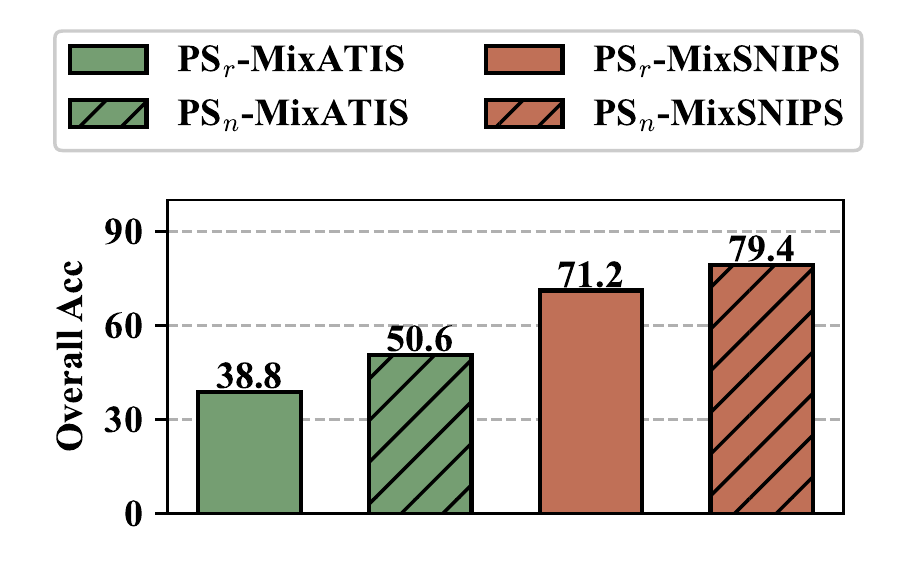}

\caption{Comparison between two strategies. We use PS$_{r}$ and PS$_{n}$ to denote {\modelname{}} utilizing raw labels or natural language descriptions of labels.}
\label{fig:overall} 
\end{figure}
\begin{figure}[t]
\centering
\includegraphics[width=0.38\textwidth]{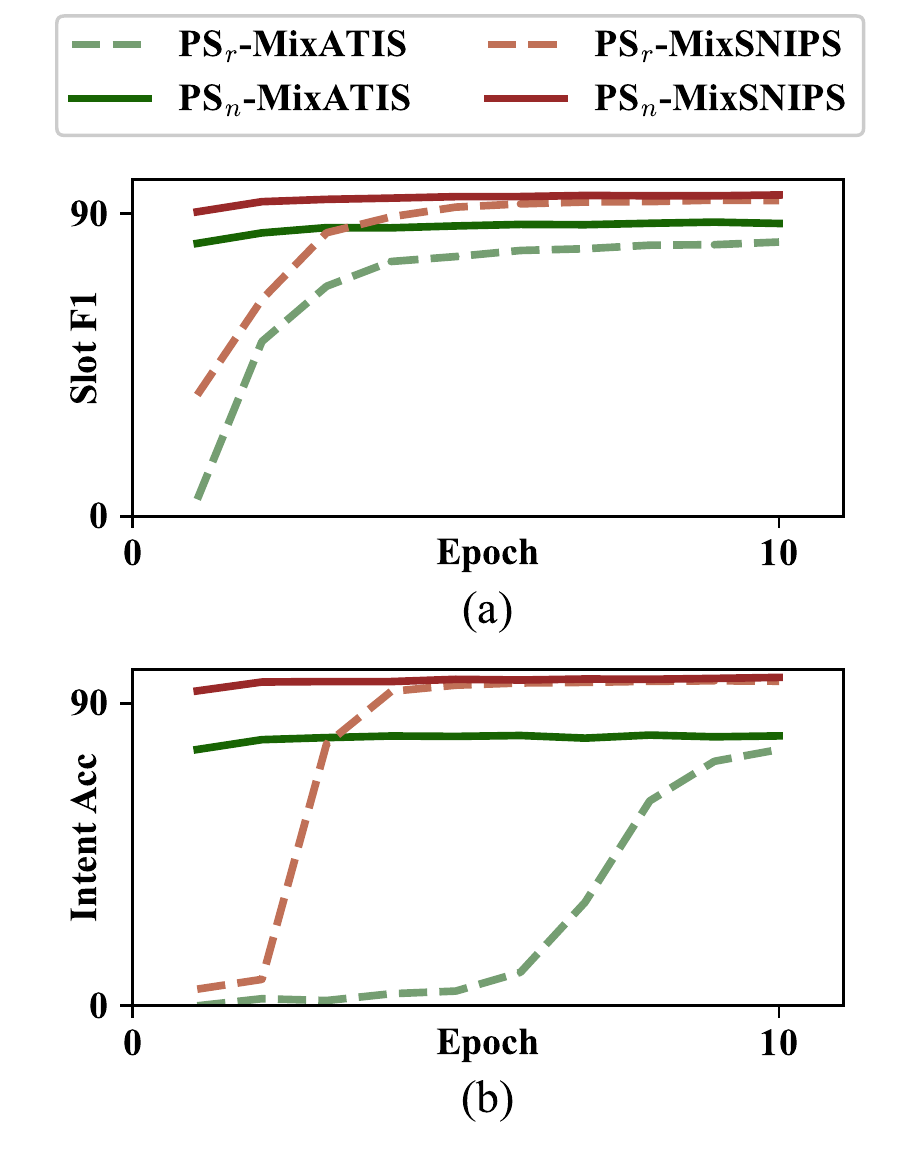}

\caption{Dynamic change of performance on Slot F1 and Intent Acc along the training process.}
\label{fig:plot} 
\end{figure}

\begin{table*}
\centering
\scalebox{0.78}{
     \begin{tabular}{lllllll}
          \toprule
          \multicolumn{1}{l}{\textbf{Input}} & \multicolumn{1}{l}{\textbf{Output}} \\ \midrule
          \multicolumn{1}{l}{} & \multicolumn{1}{l}{\begin{tabular}{l}
               \textbf{ID:} intent: \underline{\textit{atis\_airport}}, intent: \underline{\textit{atis\_flight}}
          \end{tabular}} \\ 
          \multicolumn{1}{l}{\multirow{3}{*}{\begin{tabular}{l}
               \textbf{User:} describe pittsburgh airport and then \\
               list flights from denver to san francisco no \\
               denver to philadelphia.
          \end{tabular}}} & \multicolumn{1}{l}{\begin{tabular}{l}
               \textbf{SF:} \underline{\textit{airport\_name}}: pittsburgh airport, \underline{\textit{fromloc.city\_name}}: \\
               denver, \underline{\textit{toloc.city\_name}}: 
               san francisco, \underline{\textit{toloc.city\_name}}: \\
               philadelphia
          \end{tabular}} \\ 
          \multicolumn{1}{l}{} & \multicolumn{1}{l}{\begin{tabular}{l}
               \textbf{French Translation:} décrit l'aéroport pittsburgh \\ 
               et énumère les vols depuis la city de denver à san francisco \\
               et non pas à la city de philadelphia
          \end{tabular}} \\ 
          \midrule
          \multicolumn{1}{l}{} & \multicolumn{1}{l}{\begin{tabular}{l}
               \textbf{ID:} intent: \underline{\textit{atis\_ground\_service}}, intent: \underline{\textit{atis\_ground\_fare}}
          \end{tabular}} \\ 
          \multicolumn{1}{l}{\multirow{3}{*}{\begin{tabular}{l}
               \textbf{User:} what are the costs of car rental \\
               in dallas and also ground transportation denver.\\
          \end{tabular}}} & \multicolumn{1}{l}{\begin{tabular}{l}
               \textbf{SF:} \underline{\textit{transport\_type}}: car rental, \underline{\textit{city\_name}}: dallas, \\
               \underline{\textit{city\_name}}: denver\\
          \end{tabular}} \\ 
          \multicolumn{1}{l}{} & \multicolumn{1}{l}{\begin{tabular}{l}
               \textbf{Romanian Translation:} costul maşinii închiriate în dallas \\
               şi al transportului pe teren în city.
          \end{tabular}} \\ 
          \bottomrule
     \end{tabular}
}
\caption{\label{case_study}
Case study. We use raw labels to replace corresponding natural language descriptions. \modelname{} accurately predicts all labels in the output sentences.
}
\end{table*}

\section{Discussion}
\label{sec:discussion}

\subsection{Effect of Semantic Information}
Our framework is equipped with descriptions of labels rather than raw labels, to better understand semantic information of intents and slots. To explore the effectiveness of this operation, we do experiments by replacing the descriptions with corresponding raw labels and adding them to the vocabulary as special tokens. Figure~\ref{fig:overall} illustrates the comparison between the two strategies. We use $\textrm{PS}_{n}$ to denote the former variant of \modelname{}$_{\scriptsize\textrm{small}}$ and $\textrm{PS}_{r}$ to denote that with raw labels. We set the same hyper-parameters of $\textrm{PS}_{r}$ as those in $\textrm{PS}_{n}$. 

We make the following observations:\\
1) Figure~\ref{fig:overall} shows that the absence of semantic information hinders {\modelname{}} from aligning intents with slots, which is reflected by the drop on Overall Acc. Figure~\ref{fig:plot} further demonstrates its dynamic influence on SF and ID. It indicates that semantic information is truly acquired and crucial.\\
2) Comparing the convergence rates of two strategies on Slot F1 and Intent Acc, we see that performance of $\textrm{PS}_{n}$ quickly rises to a relatively high level. However, the performance of $\textrm{PS}_{r}$ just increases step by step at early epochs, along with the modification of label representations, which is defined by local context. It suggests that pre-trained semantic features efficiently help fine-tuning. 

\subsection{Case Study}
Table~\ref{case_study} displays several typical examples. Upon receiving user utterances, {\modelname{}} completes Intent Detection and Slot Filling with the generation of key-value sequence. 

It can be noted that {\modelname{}} can predict slot-value pairs along the order of their appearance in the user input, which may be further utilized for supplementary requirements. Moreover, there exist slot-value pairs with sharing slots but distinct values, which pose a great challenge to the comprehension of networks. However, {\modelname{}} precisely predicts them without the lack of intrinsic order.

Surprisingly, we find that {\modelname{}} sometimes manages to translate user utterances into different language versions, with specific prefixes. This discovery suggests that {\modelname{}} may alleviate forgetting knowledge acquired in the pre-training stage after fine-tuning. We present it in Table~\ref{case_study}.

\section{Conclusion}
\label{sec:conclusion}
In this paper, we propose {\modelname{}}, which handles Intent Detection and Slot Filling in multi-intent SLU with a prompt-based text generation framework. To the best of our knowledge, this is the first work to utilize prompts for this problem. Our framework integrates these two sub-tasks into the same formulation while distinguishing them according to diverse prompts, which simplifies model structures and enhances interpretability. Moreover, based on the semantic similarity between intents and slots, predicted intents are driven to guide the process of Slot Filling, while a new auxiliary task Slot Prediction is introduced. Experimental results show its multi-dimensional superiority against baselines on three datasets.


\begin{thebibliography}{34}
\expandafter\ifx\csname natexlab\endcsname\relax\def\natexlab#1{#1}\fi

\bibitem[{Brown et~al.(2020)Brown, Mann, Ryder, Subbiah, Kaplan, Dhariwal,
        Neelakantan, Shyam, Sastry, Askell, Agarwal, Herbert-Voss, Krueger, Henighan,
        Child, Ramesh, Ziegler, Wu, Winter, Hesse, Chen, Sigler, Litwin, Gray, Chess,
        Clark, Berner, McCandlish, Radford, Sutskever, and
        Amodei}]{NEURIPS2020_1457c0d6}
Tom Brown, Benjamin Mann, Nick Ryder, Melanie Subbiah, Jared~D Kaplan, Prafulla
        Dhariwal, Arvind Neelakantan, Pranav Shyam, Girish Sastry, Amanda Askell,
        Sandhini Agarwal, Ariel Herbert-Voss, Gretchen Krueger, Tom Henighan, Rewon
        Child, Aditya Ramesh, Daniel Ziegler, Jeffrey Wu, Clemens Winter, Chris
        Hesse, Mark Chen, Eric Sigler, Mateusz Litwin, Scott Gray, Benjamin Chess,
        Jack Clark, Christopher Berner, Sam McCandlish, Alec Radford, Ilya Sutskever,
        and Dario Amodei. 2020.
\newblock \href
        {https://proceedings.neurips.cc/paper/2020/file/1457c0d6bfcb4967418bfb8ac142f64a-Paper.pdf}
        {Language models are few-shot learners}.
\newblock In \emph{Advances in Neural Information Processing Systems},
        volume~33, pages 1877--1901. Curran Associates, Inc.

\bibitem[{Cai et~al.(2022)Cai, Zhou, Mi, and Faltings}]{9747477}
Fengyu Cai, Wanhao Zhou, Fei Mi, and Boi Faltings. 2022.
\newblock \href {https://doi.org/10.1109/ICASSP43922.2022.9747477} {Slim:
        Explicit slot-intent mapping with bert for joint multi-intent detection and
        slot filling}.
\newblock In \emph{2022 IEEE International Conference on Acoustics, Speech and
        Signal Processing}, ICASSP '22, pages 7607--7611.

\bibitem[{Chen et~al.(2022)Chen, Zhou, and Zou}]{9747843}
Lisong Chen, Peilin Zhou, and Yuexian Zou. 2022.
\newblock \href {https://doi.org/10.1109/ICASSP43922.2022.9747843} {Joint
        multiple intent detection and slot filling via self-distillation}.
\newblock In \emph{2022 IEEE International Conference on Acoustics, Speech and
        Signal Processing}, ICASSP '22, pages 7612--7616.

\bibitem[{Chen et~al.(2019)Chen, Zhuo, and Wang}]{chen2019bert}
Qian Chen, Zhu Zhuo, and Wen Wang. 2019.
\newblock \href {https://doi.org/10.48550/ARXIV.1902.10909} {Bert for joint
        intent classification and slot filling}.
\newblock \emph{arXiv preprint arXiv:1902.10909}.

\bibitem[{Cui et~al.(2021)Cui, Wu, Liu, Yang, and
        Zhang}]{cui-etal-2021-template}
Leyang Cui, Yu~Wu, Jian Liu, Sen Yang, and Yue Zhang. 2021.
\newblock \href {https://doi.org/10.18653/v1/2021.findings-acl.161}
        {Template-based named entity recognition using {BART}}.
\newblock In \emph{Findings of the Association for Computational Linguistics:
        ACL-IJCNLP 2021}, pages 1835--1845, Online. Association for Computational
        Linguistics.

\bibitem[{Ding et~al.(2021)Ding, Yang, Lin, and Wang}]{ijcai2021-523}
Zeyuan Ding, Zhihao Yang, Hongfei Lin, and Jian Wang. 2021.
\newblock \href {https://doi.org/10.24963/ijcai.2021/523} {Focus on
        interaction: A novel dynamic graph model for joint multiple intent detection
        and slot filling}.
\newblock In \emph{Proceedings of the Thirtieth International Joint Conference
        on Artificial Intelligence}, IJCAI '21, pages 3801--3807. International Joint
        Conferences on Artificial Intelligence Organization.
\newblock Main Track.

\bibitem[{Gangadharaiah and
        Narayanaswamy(2019)}]{gangadharaiah-narayanaswamy-2019-joint}
Rashmi Gangadharaiah and Balakrishnan Narayanaswamy. 2019.
\newblock \href {https://doi.org/10.18653/v1/N19-1055} {Joint multiple intent
        detection and slot labeling for goal-oriented dialog}.
\newblock In \emph{Proceedings of the 2019 Conference of the North {A}merican
        Chapter of the Association for Computational Linguistics: Human Language
        Technologies, Volume 1 (Long and Short Papers)}, pages 564--569, Minneapolis,
        Minnesota. Association for Computational Linguistics.

\bibitem[{Haffner et~al.(2003)Haffner, Tur, and Wright}]{1198860}
P.~Haffner, G.~Tur, and J.H. Wright. 2003.
\newblock \href {https://doi.org/10.1109/ICASSP.2003.1198860} {Optimizing svms
        for complex call classification}.
\newblock In \emph{2003 IEEE International Conference on Acoustics, Speech, and
        Signal Processing}, volume~1 of \emph{ICASSP '03}, pages I--I.

\bibitem[{Hakkani-Tür et~al.(2016)Hakkani-Tür, Tur, Celikyilmaz, Chen, Gao,
        Deng, and Wang}]{hakkani-tr2016multi-domain}
Dilek Hakkani-Tür, Gokhan Tur, Asli Celikyilmaz, Yun-Nung~Vivian Chen,
        Jianfeng Gao, Li~Deng, and Ye-Yi Wang. 2016.
\newblock \href
        {https://www.microsoft.com/en-us/research/publication/multijoint/}
        {Multi-domain joint semantic frame parsing using bi-directional rnn-lstm}.
\newblock In \emph{Interspeech 2016-17th Annual Conference of the International
        Speech Communication Association}. ISCA.

\bibitem[{Kim et~al.(2017)Kim, Ryu, and Lee}]{onlyID}
Byeongchang Kim, Seonghan Ryu, and Gary~Geunbae Lee. 2017.
\newblock \href {https://doi.org/10.1007/s11042-016-3724-4} {Two-stage
        multi-intent detection for spoken language understanding}.
\newblock \emph{Multimedia Tools Appl.}, 76(9):11377–11390.

\bibitem[{Lei et~al.(2018)Lei, Jin, Kan, Ren, He, and
        Yin}]{lei-etal-2018-sequicity}
Wenqiang Lei, Xisen Jin, Min-Yen Kan, Zhaochun Ren, Xiangnan He, and Dawei Yin.
        2018.
\newblock \href {https://doi.org/10.18653/v1/P18-1133} {{S}equicity:
        Simplifying task-oriented dialogue systems with single sequence-to-sequence
        architectures}.
\newblock In \emph{Proceedings of the 56th Annual Meeting of the Association
        for Computational Linguistics (Volume 1: Long Papers)}, pages 1437--1447,
        Melbourne, Australia. Association for Computational Linguistics.

\bibitem[{Lester et~al.(2021)Lester, Al-Rfou, and
        Constant}]{lester-etal-2021-power}
Brian Lester, Rami Al-Rfou, and Noah Constant. 2021.
\newblock \href {https://doi.org/10.18653/v1/2021.emnlp-main.243} {The power of
        scale for parameter-efficient prompt tuning}.
\newblock In \emph{Proceedings of the 2021 Conference on Empirical Methods in
        Natural Language Processing}, pages 3045--3059, Online and Punta Cana,
        Dominican Republic. Association for Computational Linguistics.

\bibitem[{Liu et~al.(2021)Liu, Yuan, Fu, Jiang, Hayashi, and
        Neubig}]{liu2021pre}
Pengfei Liu, Weizhe Yuan, Jinlan Fu, Zhengbao Jiang, Hiroaki Hayashi, and
        Graham Neubig. 2021.
\newblock \href {https://doi.org/10.48550/ARXIV.2107.13586} {Pre-train, prompt,
        and predict: A systematic survey of prompting methods in natural language
        processing}.
\newblock \emph{arXiv preprint arXiv:2107.13586}.

\bibitem[{Liu et~al.(2019)Liu, Ott, Goyal, Du, Joshi, Chen, Levy, Lewis,
        Zettlemoyer, and Stoyanov}]{https://doi.org/10.48550/arxiv.1907.11692}
Yinhan Liu, Myle Ott, Naman Goyal, Jingfei Du, Mandar Joshi, Danqi Chen, Omer
        Levy, Mike Lewis, Luke Zettlemoyer, and Veselin Stoyanov. 2019.
\newblock \href {https://doi.org/10.48550/ARXIV.1907.11692} {Roberta: A
        robustly optimized bert pretraining approach}.
\newblock \emph{arXiv preprint arXiv:1907.11692}.

\bibitem[{Loshchilov and Hutter(2019)}]{loshchilov2018decoupled}
Ilya Loshchilov and Frank Hutter. 2019.
\newblock \href {https://openreview.net/forum?id=Bkg6RiCqY7} {Decoupled weight
        decay regularization}.
\newblock In \emph{International Conference on Learning Representations}.

\bibitem[{Paolini et~al.(2021)Paolini, Athiwaratkun, Krone, Ma, Achille,
        ANUBHAI, dos Santos, Xiang, and Soatto}]{paolini2020structured}
Giovanni Paolini, Ben Athiwaratkun, Jason Krone, Jie Ma, Alessandro Achille,
        RISHITA ANUBHAI, Cicero~Nogueira dos Santos, Bing Xiang, and Stefano Soatto.
        2021.
\newblock \href {https://openreview.net/forum?id=US-TP-xnXI} {Structured
        prediction as translation between augmented natural languages}.
\newblock In \emph{International Conference on Learning Representations}.

\bibitem[{Qin et~al.(2019)Qin, Che, Li, Wen, and Liu}]{qin-etal-2019-stack}
Libo Qin, Wanxiang Che, Yangming Li, Haoyang Wen, and Ting Liu. 2019.
\newblock \href {https://doi.org/10.18653/v1/D19-1214} {A stack-propagation
        framework with token-level intent detection for spoken language
        understanding}.
\newblock In \emph{Proceedings of the 2019 Conference on Empirical Methods in
        Natural Language Processing and the 9th International Joint Conference on
        Natural Language Processing (EMNLP-IJCNLP)}, pages 2078--2087, Hong Kong,
        China. Association for Computational Linguistics.

\bibitem[{Qin et~al.(2021)Qin, Wei, Xie, Xu, Che, and Liu}]{qin-etal-2021-gl}
Libo Qin, Fuxuan Wei, Tianbao Xie, Xiao Xu, Wanxiang Che, and Ting Liu. 2021.
\newblock \href {https://doi.org/10.18653/v1/2021.acl-long.15} {{GL}-{GIN}:
        Fast and accurate non-autoregressive model for joint multiple intent
        detection and slot filling}.
\newblock In \emph{Proceedings of the 59th Annual Meeting of the Association
        for Computational Linguistics and the 11th International Joint Conference on
        Natural Language Processing (Volume 1: Long Papers)}, pages 178--188, Online.
        Association for Computational Linguistics.

\bibitem[{Qin et~al.(2020)Qin, Xu, Che, and Liu}]{qin-etal-2020-agif}
Libo Qin, Xiao Xu, Wanxiang Che, and Ting Liu. 2020.
\newblock \href {https://doi.org/10.18653/v1/2020.findings-emnlp.163} {{AGIF}:
        An adaptive graph-interactive framework for joint multiple intent detection
        and slot filling}.
\newblock In \emph{Findings of the Association for Computational Linguistics:
        EMNLP 2020}, pages 1807--1816, Online. Association for Computational
        Linguistics.

\bibitem[{Raffel et~al.(2020)Raffel, Shazeer, Roberts, Lee, Narang, Matena,
        Zhou, Li, and Liu}]{raffel2019exploring}
Colin Raffel, Noam Shazeer, Adam Roberts, Katherine Lee, Sharan Narang, Michael
        Matena, Yanqi Zhou, Wei Li, and Peter~J. Liu. 2020.
\newblock \href {http://jmlr.org/papers/v21/20-074.html} {Exploring the limits
        of transfer learning with a unified text-to-text transformer}.
\newblock \emph{Journal of Machine Learning Research}, 21(140):1--67.

\bibitem[{Ravuri and Stolcke(2015)}]{ravuri2015recurrent}
Suman Ravuri and Andreas Stolcke. 2015.
\newblock \href {https://doi.org/10.21437/Interspeech.2015-42} {Recurrent
        neural network and lstm models for lexical utterance classification}.
\newblock In \emph{Interspeech 2015-16th Annual Conference of the International
        Speech Communication Association}, pages 135--139.

\bibitem[{Raymond and Riccardi(2007)}]{raymond2007generative}
Christian Raymond and Giuseppe Riccardi. 2007.
\newblock \href {https://doi.org/10.21437/Interspeech.2007-448} {Generative and
        discriminative algorithms for spoken language understanding}.
\newblock In \emph{Interspeech 2007-8th Annual Conference of the International
        Speech Communication Association}, pages 1605--1608.

\bibitem[{Schapire and Singer(2000)}]{schapire2000boostexter}
Robert~E Schapire and Yoram Singer. 2000.
\newblock \href {https://doi.org/10.1023/A:1007649029923} {Boostexter: A
        boosting-based system for text categorization}.
\newblock \emph{Machine learning}, 39(2):135--168.

\bibitem[{Su et~al.(2022)Su, Shu, Mansimov, Gupta, Cai, Lai, and
        Zhang}]{su-etal-2022-multi}
Yixuan Su, Lei Shu, Elman Mansimov, Arshit Gupta, Deng Cai, Yi-An Lai, and
        Yi~Zhang. 2022.
\newblock \href {https://doi.org/10.18653/v1/2022.acl-long.319} {Multi-task
        pre-training for plug-and-play task-oriented dialogue system}.
\newblock In \emph{Proceedings of the 60th Annual Meeting of the Association
        for Computational Linguistics (Volume 1: Long Papers)}, pages 4661--4676,
        Dublin, Ireland. Association for Computational Linguistics.

\bibitem[{Tassias(2021)}]{tassias2021prompting}
Panagiotis Tassias. 2021.
\newblock \href {http://nlp.cs.aueb.gr/theses/p_tassias_msc_thesis.pdf} {A
        prompting-based encoder-decoder approach to intent recognition and slot
        filling}.
\newblock Master's thesis, Athens University of Economics and Business.

\bibitem[{Tur et~al.(2011)Tur, Hakkani-Tür, Heck, and Parthasarathy}]{5947636}
Gokhan Tur, Dilek Hakkani-Tür, Larry Heck, and S.~Parthasarathy. 2011.
\newblock \href {https://doi.org/10.1109/ICASSP.2011.5947636} {Sentence
        simplification for spoken language understanding}.
\newblock In \emph{2011 IEEE International Conference on Acoustics, Speech and
        Signal Processing}, ICASSP '11, pages 5628--5631.

\bibitem[{Wang et~al.(2022{\natexlab{a}})Wang, Li, Yan, Yan, Wang, Wu, and
        Xu}]{wang2022instructionner}
Liwen Wang, Rumei Li, Yang Yan, Yuanmeng Yan, Sirui Wang, Wei Wu, and Weiran
        Xu. 2022{\natexlab{a}}.
\newblock \href {https://doi.org/10.48550/ARXIV.2203.03903} {Instructionner: A
        multi-task instruction-based generative framework for few-shot ner}.
\newblock \emph{arXiv preprint arXiv:2203.03903}.

\bibitem[{Wang et~al.(2022{\natexlab{b}})Wang, Xu, Liu, Zhou, Cao, Chang, and
        Sui}]{wang2021enhanced}
Peiyi Wang, Runxin Xu, Tianyu Liu, Qingyu Zhou, Yunbo Cao, Baobao Chang, and
        Zhifang Sui. 2022{\natexlab{b}}.
\newblock \href {https://arxiv.org/pdf/2109.13023.pdf} {An enhanced span-based
        decomposition method for few-shot sequence labeling}.
\newblock In \emph{Proceedings of the 2022 Conference of the North American
        Chapter of the Association for Computational Linguistics (NAACL).}

\bibitem[{Wolf et~al.(2020)Wolf, Debut, Sanh, Chaumond, Delangue, Moi, Cistac,
        Rault, Louf, Funtowicz, Davison, Shleifer, von Platen, Ma, Jernite, Plu, Xu,
        Le~Scao, Gugger, Drame, Lhoest, and Rush}]{wolf-etal-2020-transformers}
Thomas Wolf, Lysandre Debut, Victor Sanh, Julien Chaumond, Clement Delangue,
        Anthony Moi, Pierric Cistac, Tim Rault, Remi Louf, Morgan Funtowicz, Joe
        Davison, Sam Shleifer, Patrick von Platen, Clara Ma, Yacine Jernite, Julien
        Plu, Canwen Xu, Teven Le~Scao, Sylvain Gugger, Mariama Drame, Quentin Lhoest,
        and Alexander Rush. 2020.
\newblock \href {https://doi.org/10.18653/v1/2020.emnlp-demos.6} {Transformers:
        State-of-the-art natural language processing}.
\newblock In \emph{Proceedings of the 2020 Conference on Empirical Methods in
        Natural Language Processing: System Demonstrations}, pages 38--45, Online.
        Association for Computational Linguistics.

\bibitem[{Zhang et~al.(2019)Zhang, Li, Du, Fan, and Yu}]{zhang-etal-2019-joint}
Chenwei Zhang, Yaliang Li, Nan Du, Wei Fan, and Philip Yu. 2019.
\newblock \href {https://doi.org/10.18653/v1/P19-1519} {Joint slot filling and
        intent detection via capsule neural networks}.
\newblock In \emph{Proceedings of the 57th Annual Meeting of the Association
        for Computational Linguistics}, pages 5259--5267, Florence, Italy.
        Association for Computational Linguistics.

\bibitem[{Zhang et~al.(2021)Zhang, Shi, Shou, Gong, Wang, and
        Zeng}]{zhang2021joint}
Linhao Zhang, Yu~Shi, Linjun Shou, Ming Gong, Houfeng Wang, and Michael Zeng.
        2021.
\newblock \href {https://doi.org/10.48550/ARXIV.2107.11768} {A joint and
        domain-adaptive approach to spoken language understanding}.
\newblock \emph{arXiv preprint arXiv:2107.11768}.

\bibitem[{Zhang and Wang(2019)}]{zhang2019using}
Linhao Zhang and Houfeng Wang. 2019.
\newblock \href {https://doi.org/10.1007/978-3-030-32233-5_11} {Using
        bidirectional transformer-crf for spoken language understanding}.
\newblock In \emph{Natural Language Processing and Chinese Computing}, pages
        130--141, Cham. Springer International Publishing.

\bibitem[{Zhang and Wang(2016)}]{zhang2016joint}
Xiaodong Zhang and Houfeng Wang. 2016.
\newblock \href {https://www.ijcai.org/Proceedings/16/Papers/425.pdf} {A joint
        model of intent determination and slot filling for spoken language
        understanding}.
\newblock In \emph{Proceedings of the Twenty-Fifth International Joint
        Conference on Artificial Intelligence}, volume~16 of \emph{IJCAI '16}, pages
        2993--2999.

\bibitem[{Zhong et~al.(2021)Zhong, Lee, Zhang, and
        Klein}]{zhong-etal-2021-adapting-language}
Ruiqi Zhong, Kristy Lee, Zheng Zhang, and Dan Klein. 2021.
\newblock \href {https://doi.org/10.18653/v1/2021.findings-emnlp.244} {Adapting
        language models for zero-shot learning by meta-tuning on dataset and prompt
        collections}.
\newblock In \emph{Findings of the Association for Computational Linguistics:
        EMNLP 2021}, pages 2856--2878, Punta Cana, Dominican Republic. Association
        for Computational Linguistics.

\end{thebibliography}

\bibliographystyle{acl_natbib}

% \appendix
% \section{Example Appendix}
% \label{sec:appendix}

% This is an appendix.

\end{document}